\documentclass[journal]{IEEEtran}

\usepackage{graphicx}
\usepackage{booktabs}
\usepackage{multirow}
\usepackage{amssymb}
\usepackage{rotating}
\usepackage{amssymb}
\usepackage{amsmath}
\usepackage{amsthm}
\usepackage{mathtools}
\usepackage{bm}

\usepackage{listings}
\usepackage{xcolor}

\lstdefinestyle{json}{
    basicstyle=\ttfamily\footnotesize,
    breaklines=true,
    frame=single,
    numbers=left,
    numberstyle=\tiny,
    showstringspaces=false
}

\makeatletter
\def\footnoterule{\kern -3pt \hrule width 0.4\columnwidth \kern 2.6pt}
\makeatother

\begin{document}

\title{\textbf{CoFEE: Reasoning Control for LLM-Based Feature Discovery}}

\author{
\IEEEauthorblockN{Maximilian Westermann\IEEEauthorrefmark{1}, Ben Griffin\IEEEauthorrefmark{1}, Aaron Ontoyin Yin\IEEEauthorrefmark{2}, Zakari Salifu\IEEEauthorrefmark{2}, Yagiz Ihlamur\IEEEauthorrefmark{3}, Kelvin Amoaba\IEEEauthorrefmark{2}, Joseph Ternasky\IEEEauthorrefmark{2}, Fuat Alican\IEEEauthorrefmark{2}, Yigit Ihlamur\IEEEauthorrefmark{2}}

\IEEEauthorblockA{\IEEEauthorrefmark{1}University of Oxford}
\IEEEauthorblockA{\IEEEauthorrefmark{2}Vela Research}
\IEEEauthorblockA{\IEEEauthorrefmark{3}Amazon}
\thanks{
 This work was done outside of the current role that Yagiz Ihlamur holds at Amazon. Correspondence: Maximilian Westermann \textless maximilian.westermann@wadham.ox.ac.uk\textgreater.
}
}


\maketitle
\thispagestyle{empty}
\begin{abstract}
Feature discovery from complex unstructured data is fundamentally a reasoning problem: it requires identifying abstractions that are predictive of a target outcome while avoiding leakage, proxies, and post-outcome signals. With the introduction of ever-improving Large Language Models (LLMs), our method provides a structured method for addressing this challenge. LLMs are well suited for this task by being able to process large amounts of information, but unconstrained feature generation can lead to weak features. In this work, we study reasoning control in LLMs by inducing cognitive behaviors for improving feature discovery. We introduce CoFEE (Cognitive Feature Engineering Engine), a reasoning control framework that enforces cognitive behaviors in how the LLM reasons during feature discovery. From a machine learning perspective, these cognitive behaviors act as structured inductive biases over the space of candidate features generated by the model. These behaviors have been exploited with success in ML models, and include backward chaining from outcomes, subgoal decomposition, verification against observability and leakage criteria, and explicit backtracking of rejected reasoning paths. CoFEE does not modify model architecture, training, or inference; reasoning control is implemented entirely through structured prompts. In a controlled comparison, we show that enforcing cognitive behaviors yields features with higher empirical predictability than those under unconstrained vanilla LLM prompts. CoFEE achieves an average Success Rate Score that is 15.2\% higher than the vanilla approach, while generating 29\% fewer features and reducing costs by 53.3\%. Using held-out feature evaluation, we assess whether cognitively induced features generalize beyond the data used for discovery. Our results indicate that, in our evaluated setting, reasoning control is associated with improvements in quality and efficiency of LLM-based feature discovery.
\end{abstract}


%

\section{Introduction}

\IEEEPARstart{F}{eature} discovery is a critical step in analytical pipelines, particularly in domains that rely heavily on unstructured or complex data such as venture capital (VC). In these settings, relevant signals are often implicit, non-linear, and distributed across unstructured sources, making manual feature construction difficult.

While interpretable, rule-based models such as linear models and decision trees offer transparent, auditable reasoning, they often struggle in capturing the complexity of available data~\cite{gptree}. This leads to the inability to maximize the predictive effectiveness of features. Recent work such as GPTree addresses this limitation partially by integrating large language models (LLMs) into a decision tree, allowing transparent and interpretable decision tree reasoning to provide valuable information for VC decision-making~\cite{gptree}.

A complementary challenge, feature engineering, arises upstream of model construction. Previous work has shown that well-structured features can improve predictability across learning algorithms~\cite{feature_engineering_survey}, yet feature discovery remains a major bottleneck, especially in the extraction of unstructured data. LLMs offer automated feature discovery as a potential solution, with their ability to interpret large amounts of data using reasoning~\cite{llm_reasoning}. However, insufficient prompting of LLMs to propose features can produce features with limited predictability~\cite{prompting_limitations}.

\begin{figure}[t]
    \centering
    \includegraphics[width=\columnwidth]{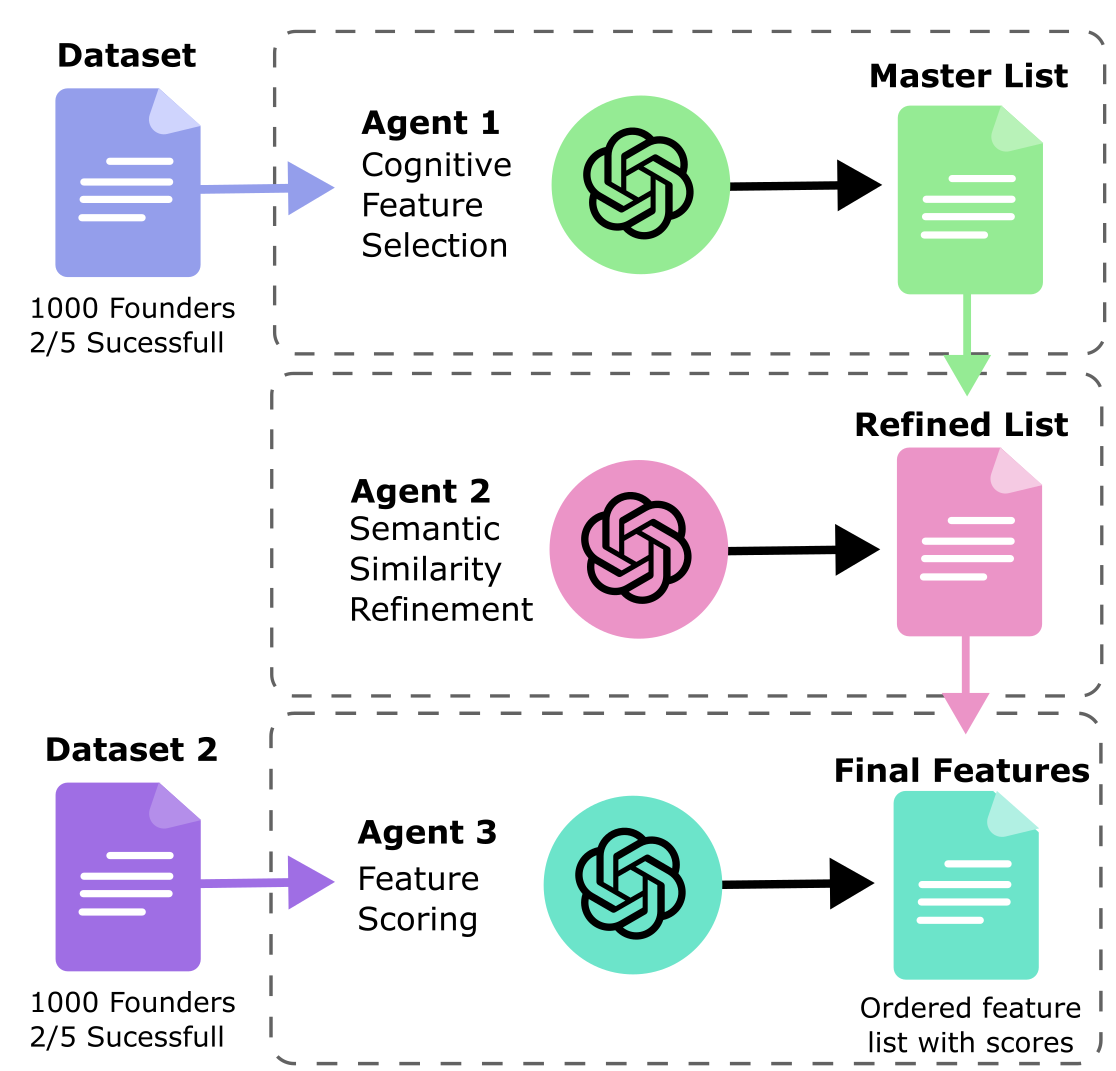}
    \caption{Overview of the CoFEE pipeline.}
\end{figure}

\begin{figure*}[t]
    \centering
    \includegraphics[width=\linewidth]{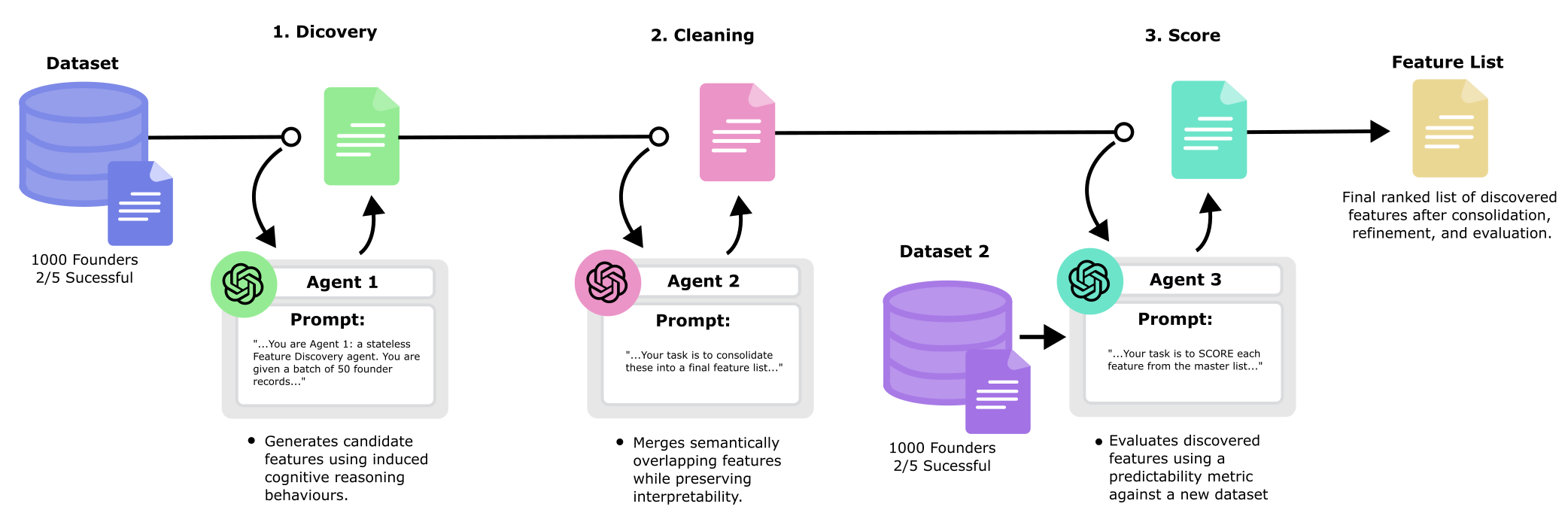}
    \caption{CoFEE pipeline. Agent 1 performs cognitive feature selection to construct an initial master list, which is refined via semantic similarity by Agent 2. Agent 3 then scores each feature by counting the number of successful and unsuccessful founders exhibiting it.}
\end{figure*}
This raises a key question: \emph{How can reasoning behaviors induced by cognitive prompting act as inductive biases to produce highly predictive and interpretable features using LLMs?} Recent work has identified a set of reasoning behaviors such as backward chaining, subgoal decomposition, verification, and backtracking that correlate with improved learning and self-improvement in language models~\cite{cognitive_llm, rl_reasoning}. Complementary research further supports this perspective, arguing that AI evaluation should mirror human testing practices by explicitly evaluating reasoning processes rather than evaluating only final outputs~\cite{ai_eval_humans}. In LLMs, such behaviors can be instantiated through prompt-level structure, motivating cognitive prompting as a practical control interface over model reasoning. Motivated by these findings, we introduce \textbf{CoFEE} (\textbf{Co}gnitive \textbf{F}eature \textbf{E}ngineering \textbf{E}ngine), an agent-based pipeline that is used to produce features (Fig. 1). Analogous to how GPTree structures LLM reasoning with decision trees, CoFEE structures LLM reasoning upstream during feature discovery. Using CoFEE, we conduct a controlled empirical study comparing the same pipeline, with one version using cognitive feature discovery carried out by an agent using cognitive prompting, and a baseline using vanilla GPT-5.2 for feature discovery.

In this work, we find that cognitive prompting consistently produces features with higher predictability scores. These findings suggest that prompting cognitive behaviors can serve as an effective design strategy for LLM-based feature engineering.

\section{Reasoning Control via Cognitive Constraints}

We use the term \textit{reasoning control} to describe how we constrain LLMs' reasoning behavior, implemented in our framework through cognitive prompting. In machine learning terms, this corresponds to imposing structured inductive biases on the feature generation process. These constraints are implemented using prompts, enforcing discipline during generation. 

In CoFEE, reasoning control is implemented through four cognitive behavior constraints as described in Gandhi et al. ~\cite{cognitive_llm}:

\begin{itemize}
    \item \textbf{Backward chaining}: Start from the desired outcome and reason backward.
    \item \textbf{Subgoal decomposition}: Break down complex tasks into smaller, manageable steps.
    \item \textbf{Verification}: Systematically check the results of each intermediate step.
    \item \textbf{Backtracking}: Revise steps explicitly when they fail. 
\end{itemize}

\section{Dataset}

To validate the effectiveness of inducing cognitive behaviors in LLMs, we use a dataset consisting of 1,000 founder profiles collected from publicly available sources. This dataset contains information about the founder and company (e.g., background, roles, sector, funding, etc.), and the success outcome. It consists of 400 (40\%) successful and 600 (60\%) unsuccessful founders. We classify 'successful' founders as those that achieve an M\&A or IPO valuation exceeding \$500M, or raised more than \$500M in total funding~\cite{vcbench}.

\subsection{Held-Out Feature Evaluation}

To assess whether these features discovered in this pipeline generalize beyond the data used for discovery, we carry out a held-out feature evaluation following standard ML practice. Rather than evaluating on the same data, we explicitly assess these features on a held-out dataset for feature evaluation.

1,000 founders are used for discovery, and we use another 1,000 founders (again with 40\% successful) to assess feature performance. The discovery set is passed through the pipeline for discovery and refinement. The held-out evaluation set is never exposed to the feature discovery pipeline.

Once discovery is completed, the features are frozen. These frozen features are then applied to the held-out evaluation set, where feature predictive quality is assessed. This ensures that no information in the held-out set influences feature discovery, allowing this evaluation to assess feature generalization.

\section{Pipeline Overview}

In this work, we present CoFEE, an agent-based pipeline that explicitly enforces structured cognitive behavior during feature discovery. While models such as GPT-5.2 already exhibit certain reasoning behaviors, previous studies suggest that these behaviors can be further strengthened and systematically induced through explicit prompting and structural constraints.
 
With CoFEE, we explore the feasibility of leveraging these enforced cognitive qualities to enhance feature discovery compared to vanilla GPT-5.2. The pipeline (see Fig. 2) breaks down this process into three specialized agents (for discovery, scoring, refining) using GPT-5.2 for each agent in this pipeline.

Viewed as an ML system, CoFEE implements a generator–evaluator loop in which cognitively constrained reasoning guides feature hypothesis generation prior to evaluation.

\subsection{Agent 1: Feature Discovery}

Agent 1 is the primary agent for this pipeline and is responsible for proposing candidate features based on the provided dataset. The agent receives structured prompts that explicitly induce cognitive behaviors, including backward chaining, subgoal decomposition, verification, and backtracking. Backward chaining is used to reason from the target outcome to features; subgoal decomposition structures feature discovery around high-level causal categories; verification enforces observability and non-proxy constraints; and backtracking records and rejects invalid reasoning paths. In the vanilla GPT-5.2, Agent 1 proposes features without these cognitive constraints.

This agent processes the dataset 50 founders at a time, extracting features and adding them to a master list until the entire dataset is analyzed. 

The initial prompts for Agent 1 of CoFEE and vanilla prompting appear below. Full prompts are provided in the Supplement.

\paragraph{CoFEE Prompt (Agent 1)}

\begin{quote}
\ttfamily
"You are Agent 1: a stateless Feature Discovery agent.

You are given a batch of 50 founder records.

You do NOT know which founders are successful or unsuccessful.
You have NO memory of previous batches.

Your task is to propose candidate FEATURES that could plausibly distinguish
successful from unsuccessful founders.

You are performing Cognitive Feature Reasoning.
You must explicitly apply the following cognitive behaviors.
You must produce structured outputs and make explicit decisions.

--------------------------------
1. BACKWARD CHAINING
--------------------------------
Start from system-level success or failure.

For each proposed mechanism:
- State the causal hypothesis explicitly.
- Explain why this mechanism would operate *before* success.
- Map the mechanism to at least one measurable or inferable quantity
  available in the dataset.

You may reason about hidden variables, but the final feature MUST be:
- observable pre-success
- expressible in deterministic logic

If a mechanism cannot be mapped to an observable feature, abandon it.

--------------------------------
2. SUBGOAL SETTING
--------------------------------
Organize exploration into NO MORE THAN 4 subgoals chosen from:
- founder capability formation
- team coordination and complementarity
- market structure and constraints
- early execution dynamics

For each subgoal:
- List candidate mechanisms.
- Maintain the hierarchy:
    system behavior → mechanism → feature.

If a subgoal:
- collapses into a proxy
- fails observability
- has ambiguous causal direction

then explicitly ABANDON or REVISE the subgoal and explain why.

--------------------------------
3. VERIFICATION
--------------------------------
For each proposed feature, verify:
- it is observable before the success outcome
- it encodes a plausible causal mechanism
- it is not a prestige-based, descriptive, or post-outcome proxy

For each feature, list:
- potential bias sources
- uncertainty or ambiguity

If verification fails, reject the feature.

--------------------------------
4. BACKTRACKING
--------------------------------
Explicitly record every abandoned reasoning path.

For each abandoned path, record:
- why it initially seemed promising
- which constraint caused rejection
  (proxy risk, leakage, observability failure, causal ambiguity)

Use these abandoned paths to bias future exploration away
from similar dead ends.

--------------------------------
\end{quote}

\paragraph{Vanilla Prompt (Agent 1)}

\begin{quote}
\ttfamily
"You are Agent 1: a stateless Feature Discovery agent.

You are given a batch of 50 founder records.

You do NOT know which founders are successful or unsuccessful.
You have NO memory of previous batches.

Your task is to propose candidate FEATURES that could plausibly distinguish
successful from unsuccessful founders.

Features must be observable pre-success

You must produce structured outputs and make explicit decisions."
\end{quote}

\subsection{Agent 2: Feature Consolidation}

Agent 2 identifies semantically overlapping features and merges those that represent the same underlying mechanism as demonstrated in Fig. 3. Merging decisions are conservative and preserve feature provenance to maintain interpretability. This step reduces redundancy while ensuring that distinct causal mechanisms remain separate.

\begin{figure}[h]
    \centering
    \includegraphics[width=0.7\columnwidth]{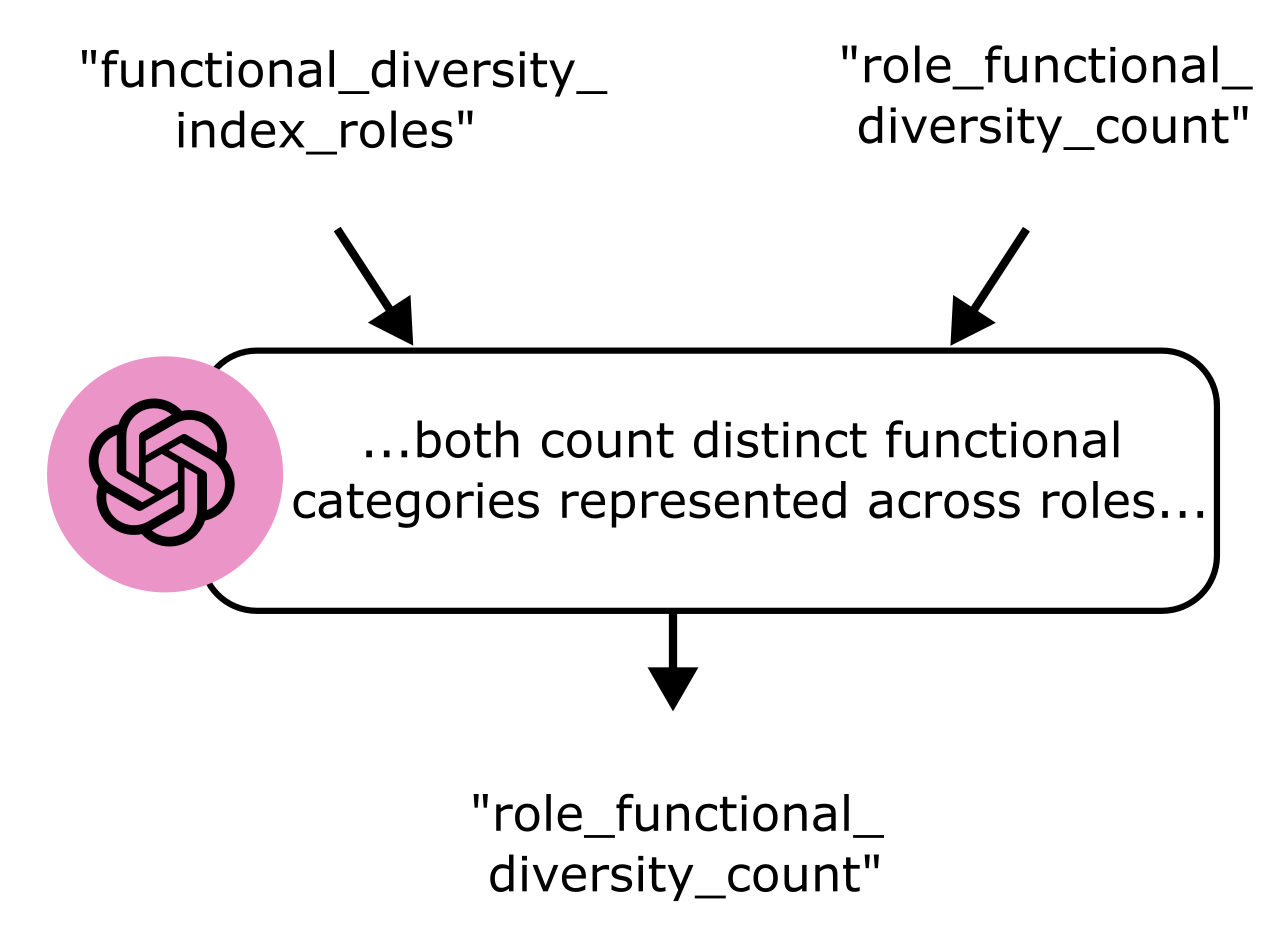}
    \caption{Diagram illustrating the Agent 2 process, in which semantically similar features are compared, their similarity is justified, and the features are combined.}
\end{figure}

\subsection{Agent 3: Scoring}

Agent~3 evaluates features by comparing feature name and definitions against founder records in batches of up to 100 features and 1{,}000 founders at a time. When a feature matches a founder, the founder is tagged with that feature. After all founders have been tagged, Agent~3 outputs a JSON file recording feature assignments.

This output is then used to deterministically compute feature statistics. $n_1$ denoting the number of successful founders exhibiting a given feature, and $n_0$ denoting the number of unsuccessful founders exhibiting the feature. We define the success-rate delta ($\Delta SR$) as the difference between the success probability conditioned on feature presence and the success probability conditioned on feature absence:

\begin{equation}
\Delta SR
=
\frac{n_1}{n_1 + n_0}
-
\frac{(N_1 - n_1)}{(N_1 - n_1) + (N_0 - n_0)}
\end{equation}

where $N_1$ and $N_0$ denote the total numbers of successful and unsuccessful founders in the dataset, respectively.

\section{Experiment}

We evaluate the impact of cognitively structured prompting on feature discovery using a controlled experimental setup.

We compare two feature discovery conditions:

\begin{itemize}
    \item \textbf{Cognitive Prompting}: where Agent 1 is constrained to explicitly apply the cognitive behaviors: backward chaining, subgoal decomposition, verification, and backtracking.
    \item \textbf{Vanilla Prompting}: where Agent 1 is prompted to propose features without explicit cognitive constraints.
\end{itemize}

Between the two conditions, the remainder of the pipeline components are held constant (Agents 2–3). This ensures that any observable differences can be attributed to differences in Agent 1 prompting.

\subsection{Evaluation Metric}

For each feature, we compute a success-rate delta ($\Delta SR$), which measures the difference in success rates between founders who exhibit the feature and those who do not. This is computed by comparing the prevalence of the feature among successful founders to its prevalence among unsuccessful founders, with larger differences indicating stronger discriminative power. This metric allows us to quantify how effectively a given feature set separates successful founders from unsuccessful ones.

To evaluate whether cognitive prompting improves feature quality, we compare the top ten features ranked by $\Delta SR$ for each experimental condition. By examining differences in these top-ranked features, we assess whether cognitive prompting leads to systematically stronger or more discriminative features.

\section{Results}

\subsection{Feature Discovery Results}

 To ensure statistical stability and modeling relevance, we restrict analysis to features with marginal support $\geq 10\%$ of the sample (n $\geq$ 100). For a binomial proportion with n = 100, the maximum standard error (attained at p = 0.5) is 0.05. This threshold filters high-variance rare features that are unlikely to generalize out-of-sample and concentrates evaluation on predictors with meaningful population coverage. The top 10 features discovered using CoFEE (cognitive prompting) are shown in Table~\ref{tab:cognitive_single_updated}. The top 10 features discovered using vanilla prompting are shown in Table~\ref{tab:vanilla_single_updated}. Finally, the side-by-side comparison of features discovered by CoFEE and vanilla prompting are shown in Table~\ref{tab:coffee_comparison}. Further metric comparisons are presented in the Appendix.

\begin{table}[h!]
\centering
\caption{Top CoFEE features ranked by success-rate delta ($\Delta SR$)}
\label{tab:cognitive_single_updated}
\begin{tabular}{lccc}
\hline
Feature ID & $n_1$ & $n_0$ & $\Delta SR$ \\
\hline
{\scriptsize\texttt{top\_university\_education\_flag}} & 82  & 37  & 0.328 \\
{\scriptsize\texttt{education\_top10\_qs\_flag}} & 67  & 33  & 0.300 \\
{\scriptsize\texttt{highest\_degree\_level}} & 304 & 292 & 0.272 \\
{\scriptsize\texttt{education\_top50\_qs\_flag}} & 70  & 41  & 0.259 \\
{\scriptsize\texttt{technical\_background\_flag}} & 190 & 155 & 0.230 \\
{\scriptsize\texttt{functional\_role\_diversity}} & 241 & 222 & 0.224 \\
{\scriptsize\texttt{job\_count\_total}} & 254 & 241 & 0.224 \\
{\scriptsize\texttt{job\_tenure\_longest\_bucket}} & 284 & 289 & 0.224 \\
{\scriptsize\texttt{cross\_industry\_breadth\_count}} & 252 & 239 & 0.222 \\
{\scriptsize\texttt{functional\_breadth\_score}} & 245 & 229 & 0.222 \\
\hline
\end{tabular}
\end{table}

\begin{table}[h!]
\centering
\caption{Top Vanilla features ranked by success-rate delta ($\Delta SR$)}
\label{tab:vanilla_single_updated}
\begin{tabular}{lccc}
\hline
Feature ID & $n_1$ & $n_0$ & $\Delta SR$ \\
\hline
{\scriptsize\texttt{has\_top10\_qs\_education}} & 81  & 34  & 0.344 \\
{\scriptsize\texttt{data\_completeness\_score}} & 281 & 277 & 0.234 \\
{\scriptsize\texttt{max\_role\_seniority\_level}} & 269 & 269 & 0.216 \\
{\scriptsize\texttt{education\_qs\_rank\_best\_numeric}} & 217 & 201 & 0.205 \\
{\scriptsize\texttt{has\_senior\_executive\_role}} & 209 & 192 & 0.202 \\
{\scriptsize\texttt{functional\_background\_primary}} & 282 & 302 & 0.199 \\
{\scriptsize\texttt{role\_seniority\_score\_max}} & 262 & 273 & 0.193 \\
{\scriptsize\texttt{founder\_tenure\_years\_estimate}} & 144 & 122 & 0.193 \\
{\scriptsize\texttt{max\_seniority\_level}} & 263 & 275 & 0.192 \\
{\scriptsize\texttt{has\_any\_industry\_match\_prior\_job\_flag}} & 116 & 97  & 0.184 \\
\hline
\end{tabular}
\end{table}

\subsection{Feature Discovery Comparison and Interpretation}

\begin{table}[t]
\centering
\caption{Comparison of feature quality and efficiency between CoFEE and Vanilla prompting. Feature quality is evaluated using success-rate delta ($\Delta SR$) on held-out data.}
\label{tab:coffee_comparison}
\begin{tabular}{lcc}
\toprule
Metric & CoFEE & Vanilla \\
\midrule
Mean $\Delta SR$ (Top-10)   & 0.250 & 0.217 \\
Median $\Delta SR$          & 0.227 & 0.204 \\
Total Features Generated    & 157   & 222   \\
Total Cost (USD)            & \$8.54 & \$18.29 \\
\bottomrule
\end{tabular}
\end{table}

Overall, CoFEE produces a broader distribution of highly predictive features, with consistently higher $\Delta SR$ values across the top-ranked features compared to vanilla prompting.

\section{Discussion}

The results of this study suggest that explicitly structuring cognitive prompts in an LLM, specifically GPT-5.2, for feature discovery has a measurable impact on the quality of the resulting features. When cognitive behaviors such as backward chaining, subgoal decomposition, verification, and backtracking are enforced, the discovered features achieve higher predictability scores than those generated via vanilla prompting. Because all other pipeline components are held constant, the observed improvements are consistent with the presence of cognitive structure in feature discovery. This indicates that cognitive prompting provides a structured constraint on LLM feature generation behavior.

Comparing costs across the two approaches, the CoFEE evaluation incurred a total cost of \$8.54, whereas the vanilla evaluation cost \$18.29. The higher cost of the vanilla approach is attributable to the larger number of features it produced (222 features), compared to CoFEE’s 157 features. The increased feature count in the vanilla setting resulted in a greater number of API calls during the feature scoring and merging stages.

These results suggest that inducing cognitive behaviors via prompting functions as an effective inductive bias in LLM-based feature discovery systems.

Despite the empirical improvements observed, important limitations remain. First, the results are evaluated on a single domain, and do not establish generalization beyond this setting. Second, $\Delta SR$ captures empirical predictability differences but does not directly assess downstream task performance. Third, the induced cognitive behavior prompts and the robustness of these effects across alternative models, prompting formulations, and scales have not yet been evaluated. 

Future work will evaluate CoFEE across multiple domains, measure downstream model performance when incorporating discovered features, and examine robustness across different model architectures and prompt structures.

\section{Conclusion}

CoFEE provides empirical evidence that explicit reasoning control can improve LLM-based feature discovery while reducing costs by 53.3\% in our evaluated setting. By inducing structured cognitive behaviors, the pipeline produces features with higher $\Delta SR$ values obtained at lower computational cost relative to the vanilla prompt baseline. These findings suggest that reasoning control may serve as a practical pipeline design strategy for LLM-based analytical systems.

\ifCLASSOPTIONcaptionsoff
  \newpage
\fi

\bibliographystyle{IEEEtran}
\bibliography{references}

\section{Appendix}

Two tables included comparing feature quality metrics for the two discovery approaches. Precision denotes the conditional success probability among founders exhibiting the feature, $P(Y=1 \mid f=1)$. Support represents the proportion of the dataset exhibiting the feature, $(n_1+n_0)/(N_1+N_0)$, reflecting population coverage. Lift measures relative improvement over the baseline success rate and is defined as $\text{Precision} / 0.4$, where $0.4$ is the global base rate of success in the dataset.

\begin{table}[h!]
\centering
\caption{Top CoFEE features with extended metrics}
\label{tab:cognitive_extended}
\begin{tabular}{lcccccc}
\hline
Feature & $n_1$ & $n_0$ & Precision & $\Delta SR$ & Lift & Support \\
\hline
F1 & 82  & 37  & 0.689 & 0.328 & 1.723 & 0.119 \\
F2 & 67  & 33  & 0.670 & 0.300 & 1.675 & 0.100 \\
F3 & 304 & 292 & 0.510 & 0.272 & 1.275 & 0.596 \\
F4 & 70  & 41  & 0.631 & 0.259 & 1.577 & 0.111 \\
F5 & 190 & 155 & 0.551 & 0.230 & 1.377 & 0.345 \\
F6 & 241 & 222 & 0.521 & 0.224 & 1.301 & 0.463 \\
F7 & 254 & 241 & 0.513 & 0.224 & 1.283 & 0.495 \\
F8 & 284 & 289 & 0.496 & 0.224 & 1.239 & 0.573 \\
F9 & 252 & 239 & 0.513 & 0.222 & 1.283 & 0.491 \\
F10 & 245 & 229 & 0.517 & 0.222 & 1.292 & 0.474 \\
\hline
\end{tabular}
\end{table}

\begin{table}[h!]
\centering
\caption{Top Vanilla features with extended metrics}
\label{tab:vanilla_extended}
\begin{tabular}{lcccccc}
\hline
Feature & $n_1$ & $n_0$ & Precision & $\Delta SR$ & Lift & Support \\
\hline
F1 & 81  & 34  & 0.704 & 0.344 & 1.761 & 0.115 \\
F2 & 281 & 277 & 0.504 & 0.234 & 1.259 & 0.558 \\
F3 & 269 & 269 & 0.500 & 0.216 & 1.250 & 0.538 \\
F4 & 217 & 201 & 0.519 & 0.205 & 1.298 & 0.418 \\
F5 & 209 & 192 & 0.521 & 0.202 & 1.303 & 0.401 \\
F6 & 282 & 302 & 0.483 & 0.199 & 1.207 & 0.584 \\
F7 & 262 & 273 & 0.490 & 0.193 & 1.224 & 0.535 \\
F8 & 144 & 122 & 0.541 & 0.193 & 1.353 & 0.266 \\
F9 & 263 & 275 & 0.489 & 0.192 & 1.222 & 0.538 \\
F10 & 116 & 97  & 0.545 & 0.184 & 1.362 & 0.213 \\
\hline
\end{tabular}
\end{table}

\section{Supplementary}

To ensure reproducibility and structured evaluation of feature discovery outputs, we provide the exact JSON schema used by Agent 1. This template captures not only the final feature specification, but also the intermediate reasoning structure (subgoals, causal mechanisms, and abandoned ideas), enabling inspection of reasoning control effects.

\begin{lstlisting}[style=json]
{
  "batch_id": "string",
  "features": [
    {
      "feature_name": "string",
      "subgoal": "string",
      "causal_mechanism": "string",
      "definition": "string",
      "computation_logic": "string",
      "abandoned_ideas": [
        {
          "idea": "string",
          "reason": "string"
        }
      ]
    }
  ]
}



\end{lstlisting}

\end{document}